\documentclass[twoside]{article}
\usepackage[accepted]{aistats2e}
\usepackage{amsmath, xspace, amssymb, graphicx}  

\usepackage{color}
\usepackage{natbib}
\usepackage[pdftex,
            bookmarks=true,
            bookmarksnumbered=false,  
            bookmarksopen=false,      
            colorlinks=true,
            linkcolor=webred,
            pdfauthor={Nikolai Slavov},
            pdftitle={RCweb},
            pdfkeywords={Nikolai, Slavov, network, inference, RCweb},
             ]{hyperref}

\definecolor{webgreen}{rgb}{0, 0.5, 0} 
\definecolor{webblue}{rgb}{0, 0, 0.5}  
\definecolor{webred}{rgb}{0.5, 0, 0}   

\definecolor{citecolor}{rgb}{0.071, 0.36, 0.67}   
\definecolor{linkcolor}{rgb}{0.071, 0.4, 0.67}  
\hypersetup{
colorlinks=true, 
citecolor=citecolor,
linkcolor=linkcolor
}

\newcommand{\ith}{\ensuremath{i^{th} }\xspace	} 
 
\newcommand{\kth}{\ensuremath{k^{th} }\xspace	} 

\newcommand{\B}  {\ensuremath{\mathbf{B}}\xspace	}
\newcommand{\C}  {\ensuremath{\mathbf{C}}\xspace	} 
\newcommand{\D}  {\ensuremath{\mathbf{D}}\xspace	}  
\newcommand{\G}  {\ensuremath{\mathbf{G}}\xspace	}

\newcommand{\R}  {\ensuremath{\mathbf{R}}\xspace	}

\newcommand{\Q}  {\ensuremath{\mathbf{Q}}\xspace	}

\newcommand{\I}  {\ensuremath{\mathcal{I}}\xspace	} 
\newcommand{\N}  {\ensuremath{\mathcal{N}}\xspace	}
\newcommand{\cR}  {\ensuremath{\mathcal{R}}\xspace	}  

\newcommand{\hB} {\ensuremath{\mathbf{\hat{B}}}\xspace	}  
\newcommand{\hC} {\ensuremath{\mathbf{\hat{C}}}\xspace	} 
\newcommand{\hD} {\ensuremath{\mathbf{\hat{D}}}\xspace	} 
\newcommand{\hG} {\ensuremath{\mathbf{\hat{G}}}\xspace	} 
\newcommand{\hR} {\ensuremath{\mathbf{\hat{R}}}\xspace	}

\newcommand{\tG} {\ensuremath{\mathbf{ {G}}}\xspace	} 
\newcommand{\tR} {\ensuremath{\mathbf{ {R}}}\xspace	}
\newcommand{\tS} {\ensuremath{\mathbf{ {S}}}\xspace	}
\newcommand{\tU} {\ensuremath{\mathbf{ {U}}}\xspace	}
\newcommand{\tV} {\ensuremath{\mathbf{ {V}}}\xspace	}

\newcommand{\Vz} {\ensuremath{\mathbf{ {V}}_{\!\! \omega_0}  }\xspace	}

\newcommand{\iKz} {\ensuremath{\mathbf{ {K}}_{\omega_0}^{-1}  }\xspace }

\newcommand{\name}  {$\mathcal{RC}\emph{web}$\xspace	}
\newcommand{\nm}  {$\mathcal{RC}\emph{web}$\xspace	}

\usepackage[fontsize=9.85]{scrextend}
\begin{document}

\twocolumn[
\aistatstitle{ Inference of Sparse Networks with Unobserved Variables. Application to Gene Regulatory Networks }
\aistatsauthor{\textbf{Nikolai Slavov} }
\aistatsaddress{ Princeton University, NJ 08544 \\
\href{mailto: nslavov@alum.mit.edu}{ \texttt{nslavov@alum.mit.edu} }  } 
]


\begin{abstract}
Networks are a unifying framework for modeling complex systems and network inference problems are frequently encountered in many fields. Here, I develop and apply a generative approach to network inference  (\name) for the case when the network is sparse and the latent (not observed) variables affect the observed ones. From all possible factor analysis (FA) decompositions explaining the variance in the data, \name selects the FA decomposition that is consistent with a sparse underlying network. The sparsity constraint is imposed by a novel method that significantly outperforms (in terms of accuracy, robustness to noise, complexity scaling and computational efficiency) Bayesian methods and \emph{MLE} methods using $\ell1$ norm relaxation such as K--SVD and $\ell 1$--based sparse principle component analysis  (PCA). Results from simulated models demonstrate that \name  recovers exactly the model structures for sparsity as low (as non-sparse) as 50\% and with ratio of unobserved to observed variables as high as 2. \name is robust to noise, with gradual decrease in the parameter ranges as the noise level increases. 
\end{abstract}

\section{Introduction }
Factor analysis (FA) decompositions are useful for explaining the variance of observed variables in terms of fewer unobserved variables that may capture systematic effects and allow for low dimensional representation of the data. Yet, the interpretation of latent variables inferred by FA is fraught with problems. In fact, interpretation is not always expected and intended since FA may not have an underlying generative model. A prime difficulty with interpretation arises from the fact that any rotation of the factors and their loadings by an orthogonal matrix results in a different FA decomposition that explains the variance in the observed variables just as well. Therefore, in the absence of  additional information on the latent variables and their loading, FA cannot identify a unique decomposition, much less generative relationships between latent factors and observed variables. 

A frequent choice for a constraint implemented by principle component analysis (PCA) and resulting in a unique solution is that the factors are the singular vectors (and thus orthogonal to each other) of the data matrix ordered in descending order of their corresponding singular values. Yet, this choice is often motivated by computational convenience rather than by knowledge about the system that generated the data. Another type of computationally convenient constraint applied to facilitate the interpretation of  FA results is sparsity, in the form of sparse Bayesian FA \citep{West_2002, Dueck_2005, Carvalho_2008}, sparse PCA \citep{Aspremont_2007, Sigg_2008} and FA for gene regulatory networks \citep{Srebro_2001, Peer_2002}. However, papers that introduce and use sparse PCA do not consider a generative model but rather use sparsity as a convenient tool to produce interpretable factors that are linear combinations of just a few original variables. In sparse PCA, sparsity is a way to balance interpretability at the cost of slightly lower fraction of explained variance.

The algorithm described in this paper (\name) also uses a sparse prior, but \name explicitly considers the problem from a generative perspective. \name asserts that there is indeed a set of hidden variables that connect to and regulate the observed variables via a sparse network. Based on that model, I derive a network structure learning approach within explicit theoretical framework. This allows to propose an approach for sparse FA which is conceptually and computationally different from all existing approaches such as K-SVD \citep{Aharon_2005}, sparse PCA and other LARS \citep{Bradley_2004} based methods \citep{Banerjee_2007}.  \name is appropriate for analyzing  data arising from any system in which the state of each observed variable is affected by a strict subset of the unobserved variables.  To assign the inferred latent variables to physical factors, \name needs either data from perturbation experiments or prior knowledge about the factors. This framework generalizes to non-linear interactions, which is discussed elsewhere.  Furthermore, I analyze the scaling of the computational complexity of \name with the number of observed and unobserved variables, as well as the parameter space where \name can accurately infer network topologies and demonstrate its robustness to noise in the data. 

\section{Derivation }
Consider a sparse bipartite graph $\mathcal{ G = (E, N, R )}$ consisting of two sets of vertices $\mathcal N$  and $\mathcal R$ and the associated set of directed edges $\mathcal E$ connecting \cR to \N vertices. Define a graphical model in which each vertex $s$ corresponds to a random variable; $N$ observed random variables indexed by \N ($x_{\N} = \{ x_s |  s \in \mathcal N\}$) whose states  are functions of  $P$ unobserved variables indexed by $ \mathcal R$,  $x_{\cR} = \{ x_s |  s \in \mathcal R\}$. Since the states of $x_\N$ depend on (are regulated by)  $x_\cR$, I will also refer to $x_\cR$ as regulators. The functional dependencies are denoted by a set of directed edges $\mathcal E$ so that each unobserved variable $x_i | i \in  \mathcal R$ affects (and its vertex is thus connected to)  a subset of  observed variables $x_{\alpha_{i}} = \{ x_s |  s \in \alpha_{i} \subset \mathcal N\}$. Given a dataset $\G \in \mathbb R^{M \times N}$ of $M$ configurations of the observed variables $x_{\N}$, \name aims to infer the edges $\mathcal E$ and the corresponding configurations of the unobserved variables, $x_{\mathcal R}$.

 If the state of each observed variable is a linear superposition of a subset of  unobserved variables, the data \G can be modeled with a very simple \emph{generative model} (\ref{Linear}): The data is a product between $\R \in \mathbb R^{M \times P}$ (a matrix whose columns correspond to the unobserved variables and the rows correspond to the $M$ measured  configurations) and $\C \in \mathbb R^{P \times N}$, the weighted adjacency matrix of  $\mathcal G$. The unexplained variance in the data \G is captured  by the residual $\mathbf \Upsilon$.
 \begin{eqnarray} \label{Linear}
 \G = \R \C + \mathbf \Upsilon
 \end{eqnarray}
This decomposition of \G into a product of two matrices can be considered to be a type of factor analysis with \R being the factors and \C the loadings. Even when $P \ll M$  the decomposition of \G does not have a unique solution since  $ \R \C \equiv \R  \I \C \equiv \R \Q^T\Q \C  \equiv \R^* \C^*$ for any orthonormal matrix \Q. Thus the identification of a unique decomposition corresponding to the structure of $\mathcal G$ requires additional criteria constraining the decomposition. The assumption that $\mathcal G$ is sparse requires that \C  be sparse as well meaning that the state of the \ith observed vertex  $x_i |  i \in \N$ is affected by a strict subset of the unobserved variables $x_{\psi_i} = \{ x_s |  s \in \psi_i  \subset \cR\}$, the ones whose weights in the \ith column of \C are not zeros, $\C_{i\psi_i} \ne 0$ and $\C_{i\bar \psi_i } = 0$ where $\bar  \psi_i $ is the complement of $\psi_i$. Thus, to recover the structure of $\mathcal G$, \name seeks to find a decomposition of \G in which \C is sparse. The sparsity can be introduced as a regularization with a Lagrangian multiplier $\lambda$:
\begin{align} \label{obj_fun}
 ( \hC, \hR )   = \arg \min_{ \R, \C }  \|   \G - \R \C  \|_F^2  + \lambda  \|  \C \|_0 
\end{align}
In the equations above and throughout the paper $\|\C\|_F^2\!=\!\sum_{i,j} C_{ij}^2$ denotes entry-wise (Frobenius) norm, and the zero norm of a vector or matrix ($\|\C\|_0$) equals the number of non-zero elements in the array. 

To infer the network topology  \name aims to solve the optimization problem defined by (\ref{obj_fun}). Since (\ref{obj_fun}) is a NP-hard combinatorial problem, the solution can be simplified significantly by relaxing the $\ell0$ norm to $\ell1$ norm \citep{Bradley_2004}. Then the approximated problem can be tackled with interior point methods \citep{Banerjee_2007}. As an alternative approach to $\ell1$ approximation, I propose a novel method based on introducing a degree of freedom in the singular-value decomposition (SVD) of \G by inserting an invertible\footnote{\B is always invertible by construction, see section \ref{hP} and equation (\ref{a})} matrix \B. 
\begin{align}
& \tG = \tU \tS \tV^T \equiv   
	\underbrace{ \left( \tU \tS (\B^T)^{-1} \right) }_{\hR} 
	\underbrace{  \left( \B^T \tV^T  \right)   }_{\hC}		
\end{align}
The prior  (constraint) that \hC is sparse determines \B that minimizes (\ref{obj_fun}), and thus a unique decomposition. The goal of introducing \B is to reduce the combinatorial problem to one that can be solved with convex minimization.  When the factors underlying the observed variance are fewer than the observations in \G there is no need to take the full SVD; if $P$ factors are expected, only the first $P$ largest singular vectors and values from the SVD of \G are taken in that decomposition so that $\tU\tS\tV^T$ is the matrix with rank $P$ that best approximates \G in the sense of minimizing $\| \G - \tU\tS\tV^T \|_F^2$. Since conceivably sparse decompositions may use columns outside of the best $\ell 2$ approximation, \name considers taking the first $P^*$ for $P^* > P$ singular pairs. Such expanded basis is more likely to support the optimal sparse solution and especially relevant for the case when $P$ is not known. Such choice can be easily accommodated in light of the ability of \name to exclude unnecessary explanatory variables, see section \ref{hP}.

Next \name computes  \B based on the requirement  that \C is sparse for the case $N>P$. To compute \B, one may set an optimization problem (\ref{a}). Once \B is inferred, \hR and \hC can be computed easily, $\hR = \tU \tS\hB^{-1}$ and $\hC = ( \tV \hB )^T $.
\begin{align} \label{a}
&\mathbf{ \hat B} = \arg  \min_{\B} \| \tV\B \|_0, \text{ so that }  \det( \B ) > 1  
\end{align} 

The constraint on \B in (\ref{a}) is introduced to avoid trivial and degenerate solutions, such as \B being rank deficient \B.  Thus the introduction of \B reduces (\ref{obj_fun}) to a problem (\ref{a}) that is still combinatorial and might also be approximated with a more tractable problem by relaxing the $\ell0$ norm to $\ell 1$ norm and applying heuristics \citep{Cetin_2002, Candes_2007} to enhance the solution. I propose a new approach, \nm, outlined in the next section.  

\section{\nm} 
 Assume that a sparse  $\mathbf c_i^T$ corresponding to an optimal $\hat{\mathbf{b}_i}$ (the \ith column of \hB)  form $\tV \hat{\mathbf{b}_i} =  \mathbf c_i^T $ is known. Define the set of indices corresponding to non-zero elements in $\mathbf c_i^T$ with $\omega_-$ and the set of  indices corresponding to zero elements in $\mathbf c_i^T$ with $\omega_0$. Furthermore, define the matrix \Vz to be the matrix containing only the rows of \tV whose indices are in $\omega_0$.  If $\omega_0$ and thus \Vz are known, one can easily compute $\mathbf{\hat{b_i}}$ as the right singular vector of \Vz corresponding to the zero singular value. 
Since  $\mathbf{c}_i^T$ and $\omega_0$ are not known, \nm approximates $\hat{\mathbf{b}_i}$ (the smallest\footnote{By smallest singular vector I mean the singular vector corresponding to the to the smallest singular value} right singular vector of \Vz) with  $\mathbf v_{\! s}$, the smallest right singular vector of \tV. This approximation relies 
on assuming that a low rank perturbation in a matrix results in a small change in its smallest singular vectors \citep{Wilkinson_1988, Benaych-Georges_2009}. 
Thus given that \nm is looking for the sparsest solution and the set $\omega_-$ is small relative to $N$, the angle between the singular vectors of \Vz and \tV is small as well. Therefore, $\mathbf v_{\! s}$  can serve as a reasonable first approximation of $\mathbf{b}_i$. Then \nm  systematically and iteratively uses and updates $\mathbf v_{\! s}$ by  removing rows of \tV  until $\mathbf v_{\! s}$ converges to $\mathbf{b}_i$ or equivalently \Vz becomes singular for the largest set of $\omega_0$ indices. When \Vz becomes singular, all elements of $\mathbf c_i$ whose indices are in $\omega_0$ become zero. 

\nm also has an intuitive geometrical interpretation. Consider the matrix \tV mapping the unit sphere in $\mathbb R^P$ (the sphere with unit radius from  $\mathbb R^P$) to an ellipsoid in  $\mathbb R^N$. The axes of the ellipsoid are the left singular vectors of \tV. In this picture, starting with $\omega_- =   \{ \emptyset \}$ and $\omega_0 = \{1,\hdots, N\}$, solving (\ref{a}) requires moving the fewest number of indices from $\omega_0$ to $\omega_-$ so that \Vz maps a vector from  $\mathbb R^P$ to the origin. How to choose the indices to be moved? 
At each step \nm chooses $i_{|max|}$, the index of the largest element (by absolute value) of the smallest axis of the ellipsoid which is the left singular vector of \tV with the smallest singular value. \nm moves $i_{|max|}$ from $\omega_0$ to $\omega_- $  effectively selecting the dimension whose projection is easiest to eliminate and removing its largest component, which minimizes as much as possible the projection in that dimension.  \nm keeps moving indices from $\omega_0 $  to $\omega_-$ using the same procedure until the smallest right singular vector of \Vz converges to $\mathbf{b}_i$ and the smallest singular value of \Vz approaches zero. \name is guaranteed to stop after at most $(N\!-\!P\!+\!1)$ steps since after removal of $(N\!-\!P\!+\!1)$ indices from $\omega_0$, \Vz will be at most rank $P\!-\!1$. If \nm finds a sparse solution it will converge in fewer steps. 
\begin{enumerate}
\item {\bf Task: } $$ \mathbf { \hat b}_i  = \min_{ \mathbf {b_i^T b_i \ge 1} } \|\tV \mathbf b_i \|_0$$
\item  {\bf Initialization: } 
	\begin {itemize}
		\item $\omega_- =  \{ \emptyset \}$ and $\omega_0 = \{ 1, 2, \hdots, N\} $
		\item Set $\iKz = (\Vz^T \Vz)^{-1} =  \I \in \mathbb R^{P\times P}$
		\item Set $J=1$; 
		\item $i_{|max|} = \arg \max \left ( \sum_j  \left | \tV_{\!\! ij}  \right |  \right ) $ \\ $\omega_- = \{ i_{|max|} \}$, 
		$\omega_0 = \{ i | i \in \omega_0, \; i \ne i_{|max|} \}$
		\item  Update $\iKz = RankUpdate(\iKz, \tV_{\!\! i_{|max|} } )$
	\end {itemize}	
\item  {\bf Cycle: } $J=J+1$ Repeat until convergence 
	\begin {itemize}
		\item Find the eigenvector  $\mathbf v $ for \iKz  with the largest eigenvalue $\lambda_{max}$
		\item If  $\lambda_{max}^{-1} \approx 0$ or $\mathbf v^{J} \to \mathbf v^{J-1} $, $ \mathbf { \hat b}_i \equiv \mathbf v$; {\bf STOP }
		\item  Compute the left singular vector of \Vz \\ $\mathbf u  = s^{-1} \Vz  \mathbf v  $ 
		\item $i_{|max|} = \arg \max  \left[  (|u_1|, \hdots,  |u_i|, \hdots, |u_N| )  \right]$;  
		\item $\omega_- = \{ \omega_-, i_{|max|} \}$ \\ 
		$\omega_0 = \{ i | i \in \omega_0, \; i \ne i_{|max|} \}$
		\item  Update $\iKz = RankUpdate(\iKz, \tV_{\!\! i_{|max|} } )$
	\end {itemize}
\end{enumerate}
The above algorithm can compute a single vector, $\mathbf{\hat{b_i}}$, which is just one column of \hB. To find the other columns, \nm applies the same approach to the modified (inflated) matrix, which for the \ith column of \B is  $\tV^{(i)} = \tV^{(i-1)} + \tV \mathbf{ \hat{b_i}   \hat b_i^T }$ for $i = 2, \hdots, P$. Thus, after the inference of each column of \B \nm modifies  $\tV^{(i-1)}$  to  $ \tV^{(i)}$ ($\tV^{(1)} \equiv  \tV$) so that the algorithm will not replicate its choice of $\omega_0$.  Note that in the \ith update of $\tV^{(i-1)}$ the $\tV \mathbf{ \hat{b_i}   \hat b_i^T  }$ will modify only the  $\omega_-$ rows in  $\tV^{(i-1)}$ since the rows of the $\tV \mathbf{ \hat{b_i}   \hat b_i^T  }$ whose indices are in  $\omega_0$ contain only zero elements. Applying \nm to the inflated matrices avoids inferring multiple times the same $\mathbf{\hat{b_i}}$, but a $\mathbf{\hat{b_i}}$  inferred from the inflated matrix is generally going to differ at least slightly from the corresponding $\mathbf{\hat{b_i}}$ that solves (\ref{a}) for \tV. To avoid that, \nm uses the inflated matrices only for the first few iterations until  the largest (by absolute magnitude) of the Pearson correlations between the smallest eigenvector of \Vz from the current (\ith) iteration and the recovered columns of \B is less than $1-\epsilon$ and monotonically decreasing;  $\epsilon$ is chosen for numerical stability and also to reflect the similarity between the connectivity of $\mathcal R$ vertices that can be expected in the network whose topology is being recovered. A simpler alternative that works great in practice is to use the the inflated matrix for the first $k$ iterations that are enough to find a new direction for $\mathbf{\hat{b_i}}$ and then \nm switches back to \tV so that the solution is optimal for \tV. The switch requires $k$ rank update of $\iKz \equiv (\Vz^T \Vz)^{-1}$ and thus choosing $k$ small saves computations. Choosing $k$ too small, however, may not be enough to guarantee that  $\mathbf{\hat{b_i}}$  will not recapitulate a solution that is already found. Usually $k=10$ works great and can be easily increased if the new solution is very close to an old one. 

There are a few notable elements that make \nm efficient. First, \nm does almost  all computations in $\mathbb R^P$ and since $P \ll N$, $P<M$, that  saves both memory and CPU time. Second, each step requires only a few matrix-vector multiplication for computing the eigenvectors (since the change from the previous step is generally very small) and  \iKz is computed by a rank--one update which obviates matrix inversion.  

The approach that \name takes in solving (\ref{a}) does not impose specific restrictions on the distribution of the observed variables (\G), the noise in the data ($\mathbf \Upsilon$) or the latent variables \hR. However, the initial approximation of  $\mathbf{b}_i$ with  $\mathbf v_{\! s}$ can be  poor for data arising from dense networks or special worst--case datasets. As demonstrated theoretically (Benaych-Georges 2009) and tested numerically in the next section, \name performs very well at least in the absence of worst--case scenario special structures in the data.

\section{Validation} 
To evaluate the performance of \name, I first apply it to data from simulated random bipartite networks with two different topology types, (1) Erd\"{o}s \& R\`{e}nyi and (2) scale-free whose corresponding degree distributions are (1) Poisson and (2) power--law. The network topology is encoded in a weighted adjacency matrix $\C^{gold}$ and the values for the unobserved variables are drawn from a standard uniform distribution. The simulations result in data matrices $\G \in \mathbb R^{M\times N}$ containing $M$ observations of all $N$ unobserved variables. According to \name, the optimal sparse adjacency matrix (\hC) and the hidden variables (\hR) can be inferred by the decomposition, $\hG = \hR\hC$ so that \hC is as sparse as possible while \hG is as close as possible to \G. 

In addition to \nm, such decomposition can be computed by 3 classes of existing algorithms. For a comparison, I use the latest versions for which the authors report best performance: 
({\em A}) \emph{PSMF} for Bayesian matrix factorization as implemented by the author MatLab function PSMF1 \citep{Dueck_2005};  
({\em B}) \emph{BFRM 2} for Bayesian matrix factorization as implemented by the author compiled executable \citep{Carvalho_2008};  
({\em C}) \emph{emPCA} for maximum likelihood estimate (\emph{MLE}) sparse PCA \citep{Sigg_2008};
({\em D}) \emph{K-SVD} \citep{Aharon_2005}. 
 All algorithms are implemented using the code published by their authors, and with the default values of the parameters when parameters are required. The results are compared for various \emph{M, N, P}, sparsity, and noise levels. 

\subsection{Limitations}  \label{limitations}

Before comparing the results, consider some of the limitations common to all algorithms and the appropriate metrics for comparing the results. In the absence of any other information, the decomposition of \G (no matter how accurate) cannot associate hidden variables (corresponding to columns of \hR) to physical factors. Furthermore, all methods can infer \hC and \hR only up to an arbitrary diagonal scaling or permutation matrix. First consider the scaling illustrated by the following transformation by a diagonal matrix \D, $\hG = \hR\hC= \hR \I \hC = \hR(\D\D^{-1})\hC = (\hR\D)(\D^{-1}\hC) = \hR^*\hC^*$. Such transformation is going to rescale \hR to $\hR^*$ and \hC to $\hC^*$, which is just as sparse as \hC, $\|\hC^*\|_0 = \|\hC\|_0 $. Since both decompositions explain the variance in \G equally well \name (or any of the other method) cannot distinguish between them. Thus given \hC, there is a diagonal matrix \hD that scales \hC to $\C^{gold}$, the weighted adjacency matrix of $\mathcal G$. 

The second limitation is that in the absence of addition information, \name can determine \hC and \tR up to a permutation matrix. Consider comparing \hC to the adjacency matrix used in the simulations, $\C^{gold}$. Since the identity of the inferred hidden variables is not known the rows of \hC do not generally correspond to the rows of $\C^{gold}$; $\hC_i$ (the \ith row of \hC) is most likely to correspond to the $\C^{gold}$ row that is most correlated to $\hC_i$ and the Pearson correlation between the two rows quantifies the accuracy for the inference of $\hC_i$. To implement this idea, all rows of \hC and $\C^{gold}$ are first normalized to mean zero and unit variance resulting in $\C_{nor}$ and $\C_{nor}^{gold}$. The correlation matrix then is, $\mathbf \Sigma = \C_{nor}^T \C_{nor}^{gold}$ and the most likely vertex (index of the unobserved variable) corresponding to $\hC_i$ is $k = \arg \max_j ( |\Sigma_{i1}|, \hdots,  |\Sigma_{ij}|, \hdots,  |\Sigma_{iP}|)$, where $k\in \cR$. The absolute value is required because the diagonal elements of \hD can be negative. The accuracy is measured by the corresponding Pearson correlation, $\rho_i = | \Sigma_{ik}|$. An optimal solution of this matching problem can be found by using belief propagation algorithm for the simple case of a bipartite graph even the \emph{LP} relaxed version guarantees optimal solution (Sanghavi 2007). The overall accuracy is quantified by the mean correlation $\bar \rho = (1/p) \sum_{i=1}^{i=p} \rho_i$ where $p$ is the number of inferred unobserved variables and can equal to $P$ or not depending on whether the number of unobserved variables is known or not. In computing $\bar \rho$, each row of $\C^{gold}$ is allowed to correspond only to one row of \hC and vice versa. 

In addition to the two common limitations of permutation and scaling, some algorithms \citep{Aspremont_2007} for sparse PCA require $M > N$ and since this is not the case in many real world problems and in some of the datasets simulated here, those methods are not tested. Instead I chose emPCA, which does not require $M > N$ and is the latest \emph{MLE} algorithm for sparse PCA that according to its authors is more efficient than previous algorithms \citep{Sigg_2008}.

\subsection{Accuracy and complexity scaling}
\nm has a natural way for identifying the mean degree\footnote{When \nm learns all edges, the smallest singular value of \Vz approaches zero and its smallest singular vector converges to $\mathbf{\hat{b_i}}$.}. However, some of the other algorithms require the mean degree for optimal performance. To avoid underestimating an algorithm simply because it recovers networks that are too sparse or not sparse enough, I assume that the mean degree is known and it is input to all algorithms. First all algorithms are tested on a very easy inference problem, Fig.1. 
\begin{figure}[h!]
	\centering
		\includegraphics[width=0.45\textwidth]{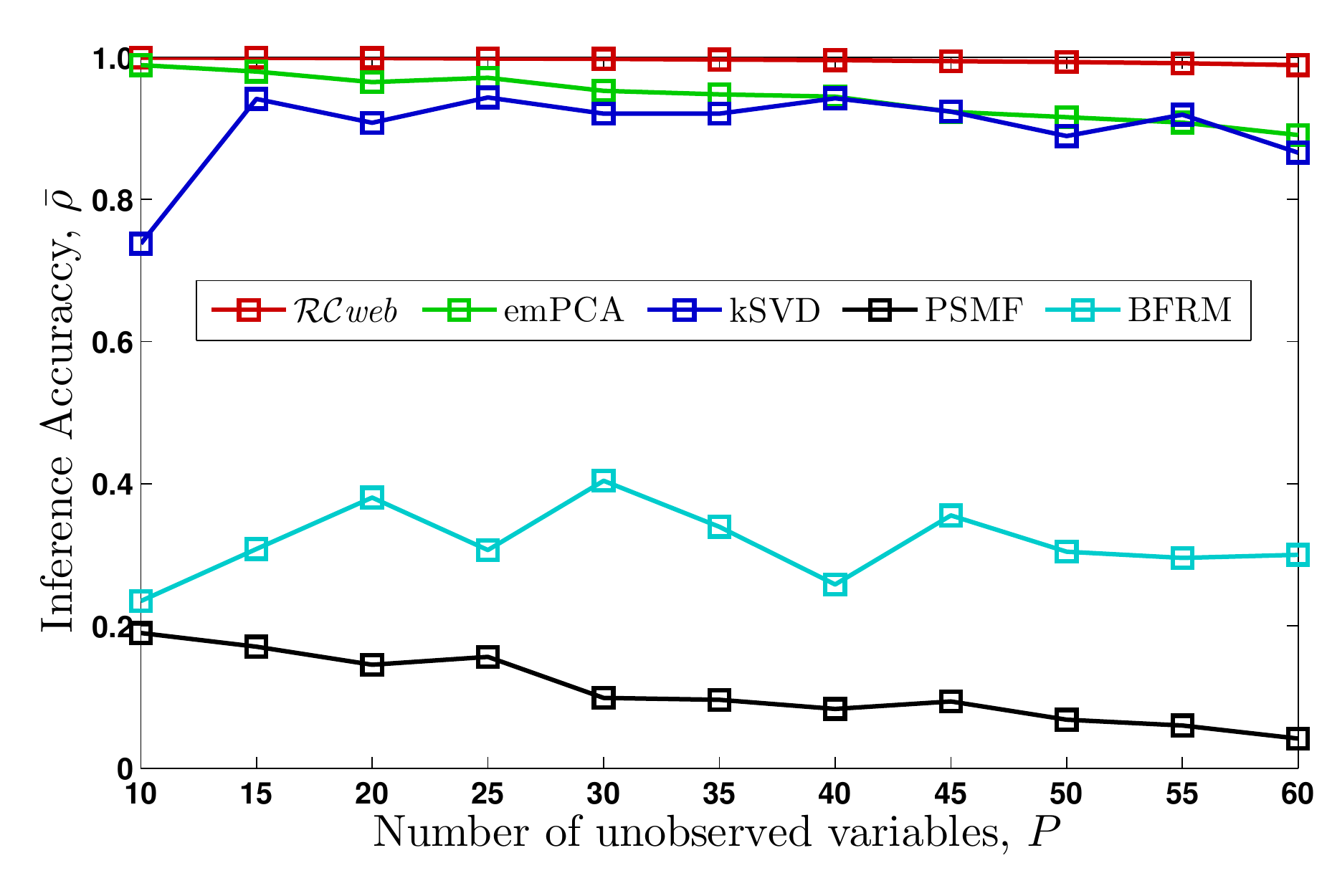}
	  \caption[Accuracy]{ Accuracy of network recovery as a function of the number of unobserved variables, $P$. Number of observed variables $N = 500$; Number of observations, $M = 280$. All networks are with Poisson mean out--degree $0.10N = 50$, with 10 \% noise in the observations.}
\end{figure}
Since PSMF and BFRM have lower accuracy and PSMF is significantly slower than the other algorithms, the rest of the results will  focus on the \emph{MLE} algorithms that also have better performance. PSMF gives less accurate results with power--law networks which can be understood in terms of the uniform prior used by PSMF. PSMF has the advantage over the \emph{MLE} algorithms in inferring a probabilistic network structure rather than a single estimate. Special advantage of BFRM is the seamless inclusion of response variables and measured factors in the inference. 
 For the \emph{MLE} algorithms, the accuracy of network inference increases with the ratio of observed to unobserved variables $N/P$ (Fig.2) and with the number of observed configurations $M$, Fig.3. In contrast, as the noise in the data and the mean out--degree (mean number of edges from \cR to \N vertices) are increased, the accuracy of the inference decreases. All algorithms perform better on Poisson networks and the lower level of noise in the data from power--law networks was chosen to partially compensate for that. 
\begin{figure}[h!]
	\centering
		\includegraphics[width=0.45\textwidth]{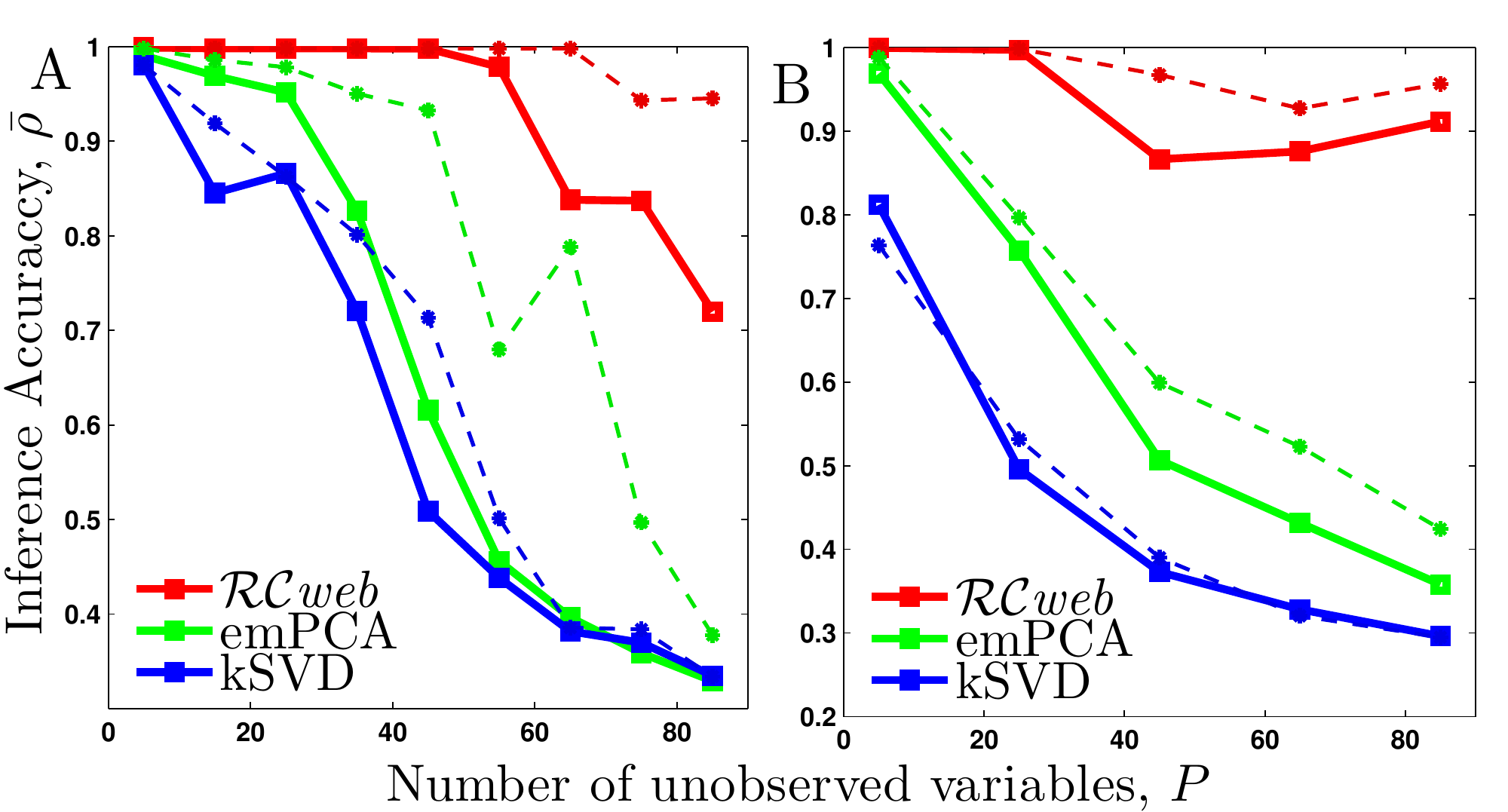}
		\includegraphics[width=0.45\textwidth]{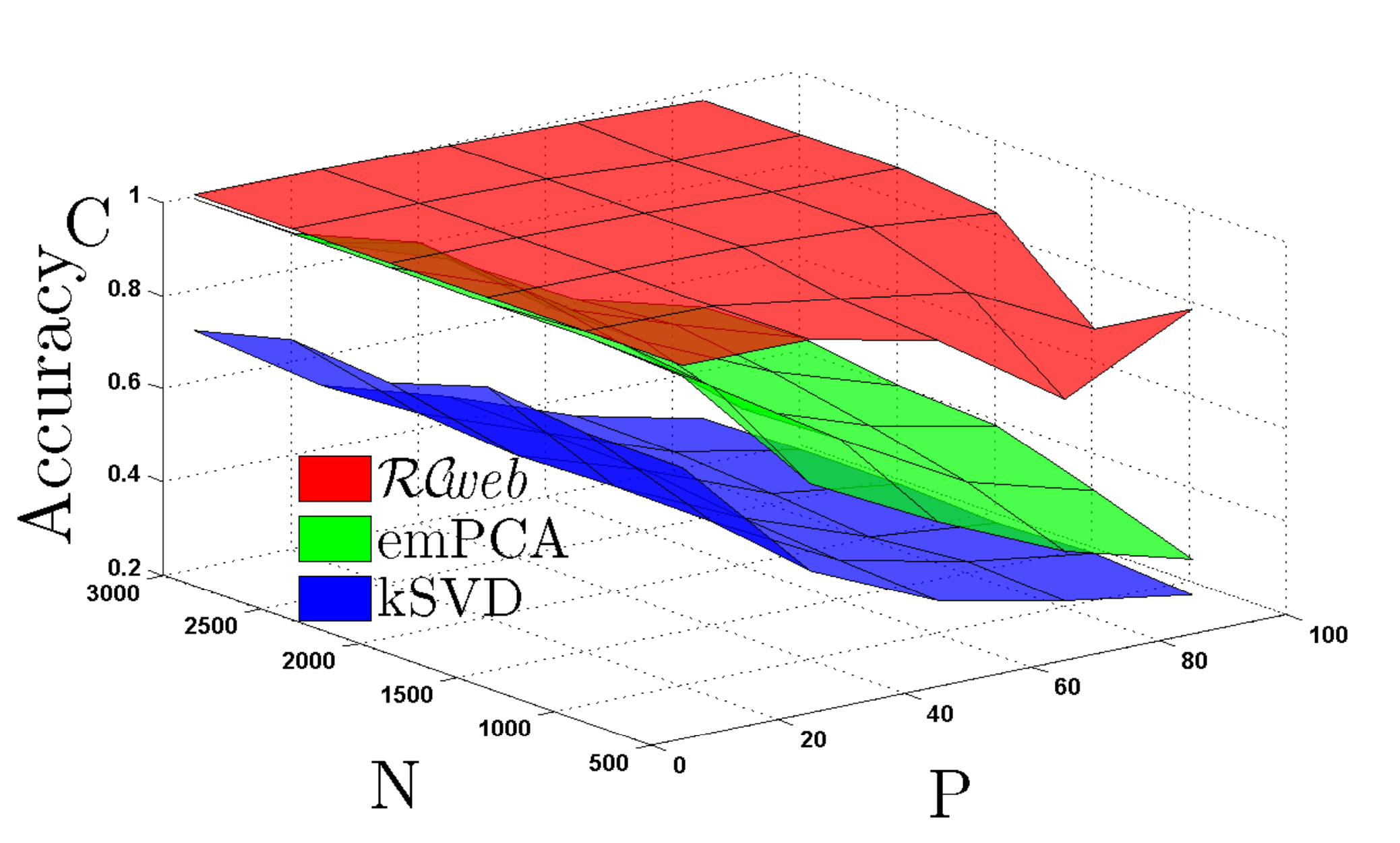}
	  \caption[Accuracy as a function of $P$]{Accuracy of network recovery as a function of the number of unobserved variables, $P$. The thicker brighter lines with squares correspond to number of observed variables $N = 500$ and the dashed, thinner lines correspond to $N = 1000$. In all cases the number of observations is $M = 2000$. A) Poisson networks with mean out--degree $0.25N$ \{125 and 250\}, with 50 \% noise in the observations and B) power--law networks (with mean out--degree $0.40N$ \{200 and 400\}, with 10 \% noise in the observations. C) The same as (B) except for wider range of the observed variables.}
\end{figure}
\begin{figure} [h]
	\centering
		\includegraphics[width=0.48\textwidth]{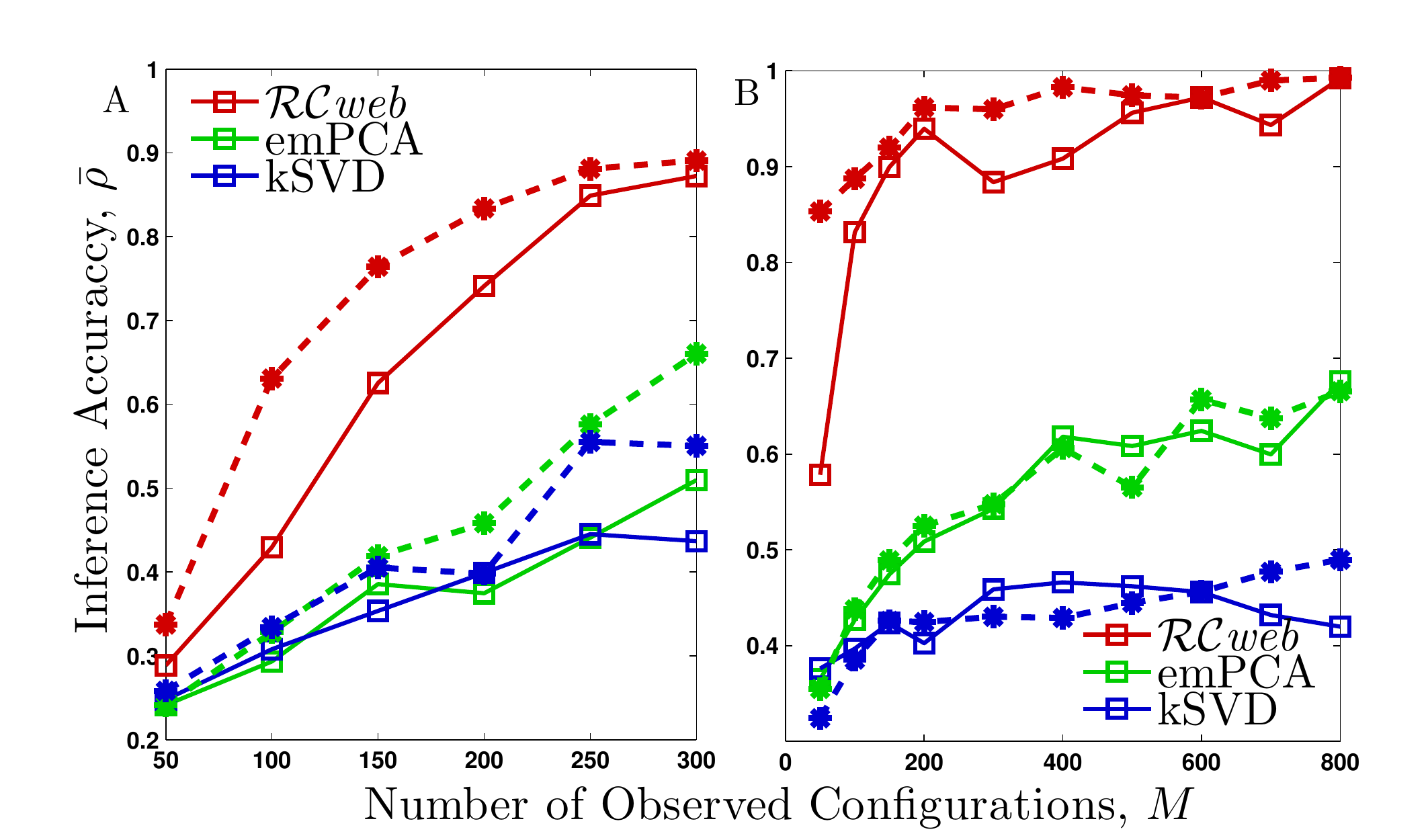}
	  \caption[Accuracy as a function of $M$]{Accuracy of network recovery as a function of the number of observed configurations, $M$. 
	   Continuous lines \& squares,  $N = 500$;  dashed lines \& circles, $N = 1000$ A) Poisson networks with mean out--degree $0.20N$ \{100 and 200\}, with 50 \% noise; B) power--law networks (with mean out--degree $0.40N$ \{200 and 400\}, with 10 \% noise. In all cases  $P = 30$.}  
\end{figure}
\begin{figure} [h!]
	\centering
		\includegraphics[width=0.48\textwidth, height=5.8cm]{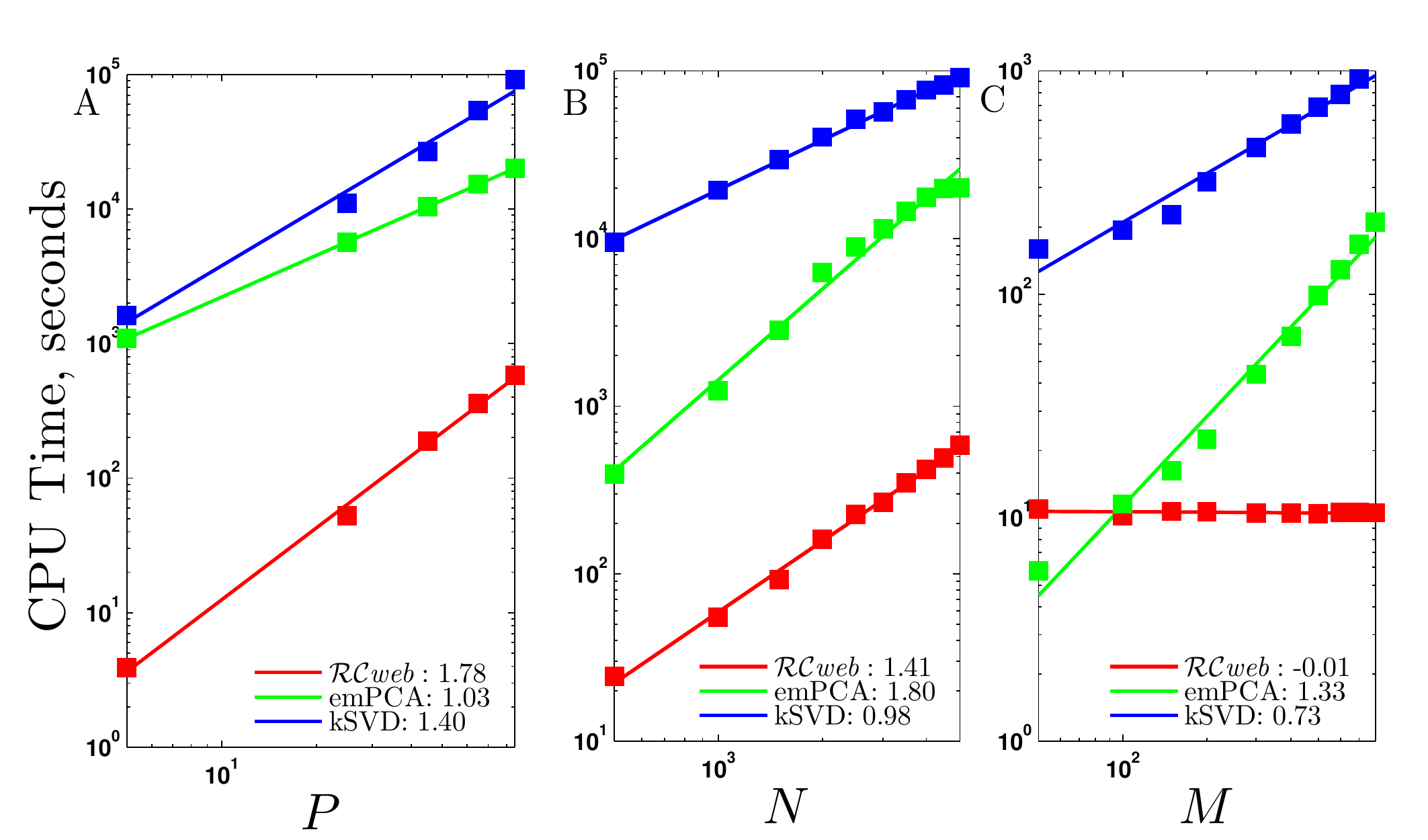}
	  \caption[Computational efficiency]{ Computational efficiency as a function of $P$, $M$ and $N$. The scaling exponents (slopes in log--log space) are reported in the legends. The networks are with power--law degree distribution and 60 \% sparsity, with mean out--degree 0.40N}
\end{figure}
An important caveat when comparing the results for different algorithms is that K-SVD iteratively improves the accuracy of the solution, and thus the output is dependent on the maximum number of iterations allowed ($I_{max}$). For the results here, $I_{max} = 20$ and the accuracy of K-SVD may improve with higher number of iterations even though I did not observe significant improvement with $I_{max} = 100$. Even at 20 iterations K-SVD is significantly slower than \name and emPCA, Fig.4.
The scaling of the algorithms with respect to a parameter was determined by holding all other parameters constant and regressing the log of the CPU time against the log of the variable parameter, Fig.4. The scaling with respect to some parameters is below the theoretical expectation since the highest complexity steps may not be speed--limiting for the ranges of parameters used in the simulations.

\subsection{Interpretability} \label{hP}
In analyzing real data the number of unobserved variables ($P$) may not be known. I observe that if a simulated network having $P$ regulators is inferred assuming $P^* > P$ regulators, the elements of \B corresponding to the excessive regulators are very close to zero, $|B_{ij}| \le  10^{-10}$ for $i\! >\! P$ or $j\! >\!P$. Thus, if the data truly originate from a sparse network \name can discard unnecessary unobserved variables.

The results of \name can be valuable even without identifying the physical factors corresponding to $x_{\cR}$. Yet identifying this correspondence, and thus overcoming the limitations outlined in section \ref{limitations} can be very desirable. One practically relevant situation allowing in-depth interpretation of the results requires measuring (if only in a few configurations) the states of some of the variables that are generally unobserved $(x_{\cR})$. This is relevant, for example, to situations in which measuring some variables is much more expensive than others (such as protein modifications versus messenger RNA concentrations) and  some $x_{s \in \cR}$ can be observed only once or a few times. 
Assume that the states of the \kth physical factor are measured (data in vector $\mathbf u_k$) in $n_k$ number of configurations, whose indices are in the set $\phi_k$. This information can be enough for determining the vertex $x_{s \in \cR}$ corresponding to the \kth factor and the corresponding $\hD_{ss}$ as follows: 1) Compute the Pearson correlations $\vec \rho$ between $\mathbf u_k$ and the columns of $\hR_{\phi_k}$. Then, the vertex of the inferred network most likely to correspond to the \kth physical factor is $s = \arg \max_i ( |\rho_{1}|, \hdots,  |\rho_{i}|, \hdots,  |\rho_{P}|)$. 2) $\hD_{ss} = (\hR_{\phi_k}^T \hR_{\phi_k})^{-1} \hR_{\phi_k}^T \mathbf u_k$. Similar to section (\ref{limitations}), weighted matching algorithms (Sanghavi 2007) can be used for finding the optimal solution if there is data for multiple $x_{s \in \cR}$. 

Partial prior knowledge about the structure of $\mathcal G$ can also be used to enhance the interpretability of \name results. Assume, for example, that some of the nodes ($x_{s \in \alpha_k \subset \mathcal N}$) regulated by the \kth physical factor (which is a hidden variable in the inference) are known. Then the matching approach that was just outlined can be used with \hC rather than \hR. If the weights are not known all non-zero elements of $\hC_{\alpha_k}$ can be set to one. The significance of the overlap (fraction of common edges) of the regulator most likely to correspond to the \kth physical and its known connectivity (coming from prior knowledge) can be quantified by a p-val (the probability of observing such overlap by chance alone) computed from the hyper--geometric distribution. This approach is exemplified with gene-expression data in the next section.

\subsection{Application to data}
Simulated models have the advantage of having known topology, and thus providing excellent basis for rigorous evaluation. Such rigorous evaluation is not possible for real biological networks. Yet, existing partial knowledge of transcriptional networks, can be combined with the approach outlined in section \ref{hP} for evaluation of inference of biological networks. In particular, consider the transcriptional network of budding yeast, Saccharomyces {\em cerevisiae}. It can be modeled as a two-layer network with transcription factors (TFs) and their complexes being regulators ($x_\cR$) whose activities cannot be measured on a high-throughput scale because of technology limitations. TFs regulate the expression levels of messenger RNAs (mRNAs) whose concentrations can be measured easily on a high-throughput scale at multiple physiological states and represent the observed variables, $x_\mathcal N$. A partial knowledge about the connectivity in this transcriptional network comes from ChIP-chip experiments \citep{MacIsaac_2006} which identify sets of genes regulated by the same transcription factors (TFs).

I first apply \name to a set mRNA configurations followed by the approach of section \ref{hP} to evaluate the results. The data matrix \G contains $423$ yeast datasets measured on the Affymetrix S98 microarray platform at a variety of physiological conditions. \name was initialized with $P^*=160$ unobserved variables (corresponding to TFs) and identified $P=153$ co-regulated sets of  genes. The inferred adjacency matrix \hC is compared directly to the adjacency matrix identified by ChIP-chip experiments \citep{MacIsaac_2006}. Indeed, sets of genes inferred by \name to be co-regulated overlap substantially with sets of genes found to be regulated by the same TF in ChIP-chip studies, Fig.5. 
\begin{figure}[h!]
	\centering
		\includegraphics[width=0.48\textwidth]{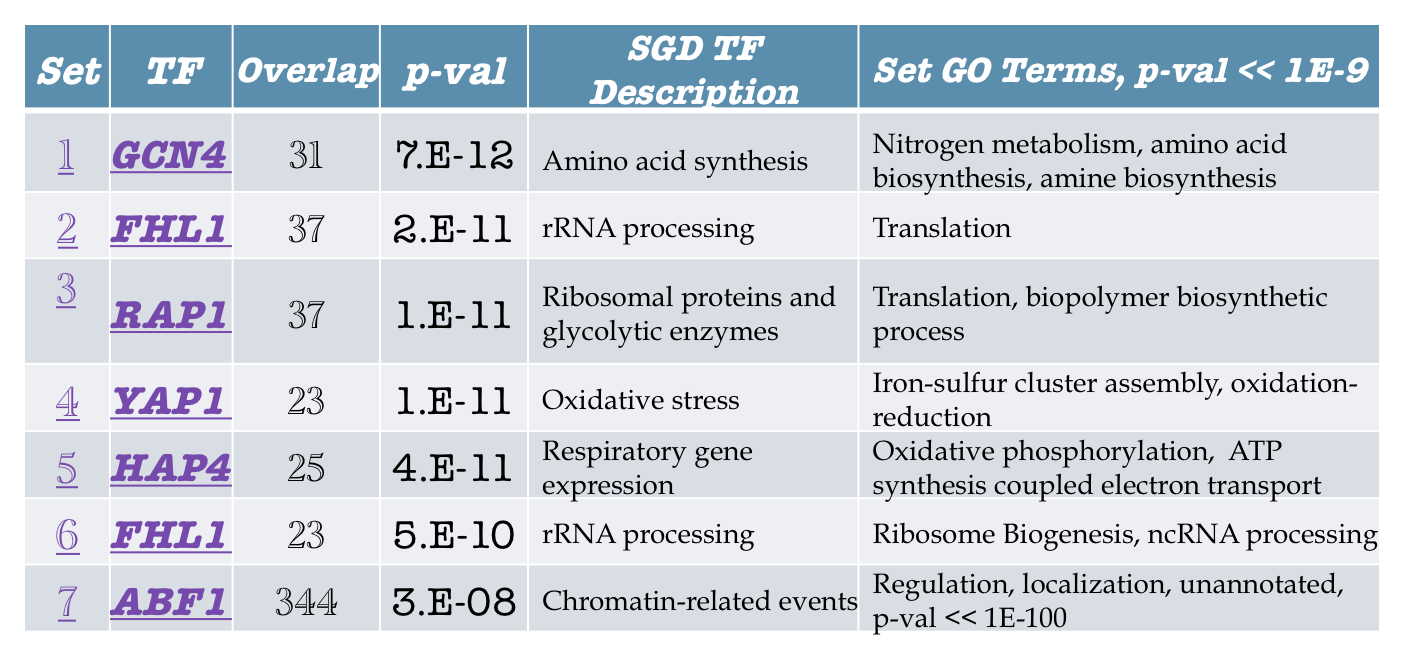}
	  \caption[Application to Data]{Overlap between the first 7 sets of genes (column 1) that \name  identifies to be regulated by a common regulator and sets of genes whose promoters are bound by the same TF (column 2) in ChIP-chip experiments. The second and third columns list the number of overlapping genes between the \name set and the ChIP-chip set and the  corresponding probabilities of observing such overlaps by chance alone.  The fourth column contains a description of the TFs from the second column and the fifth column lists the most enriched GO terms for the \name gene sets.}
\end{figure}
Furthermore, the most enriched gene ontology (GO) terms associated with the \name inferred sets of genes correspond to the functions of the respective TFs. This fact provides further support to the accuracy of the inferred network topology and suggests that many of the genes inferred to be regulated by the corresponding TFs may be {\em bona fide} targets that were not detected in the ChIP-chip experiments because of the rather limited set of physiological conditions in which the experiments are performed \citep{MacIsaac_2006}. Since the best ``gold standard'' is the only partially known topology, comparing the performance of the different algorithms in this setting can be hard to interpret; an algorithm can be penalized for correctly inferred edges if these edges are absent from the incomplete ChIP topology. An objective comparison requires experimental testing of predicted new edges and that is the subject of a forthcoming paper.   

The comparison between ChIP--chip gene sets and \name gene sets allows to identify the likely correspondences between $x_{s \in \cR}$  and TFs and to infer the activities of the TFs which are very hard to measure experimentally but crucially important to understanding the dynamics and logic of gene regulatory networks \citep{Slavov_2009,ymc_2010,slavov_thesis}. %

\section{Conclusions} 
This paper introduces an approach (\name) for inferring latent (unobserved) factors explaining the behavior of observed variables. \name aims at inferring a sparse bipartite graph in which vertices connect inferred latent factors (e.g. regulators of mRNA transcription and degradation) to observed variables (e.g. target mRNAs).  The salient difference distinguishing \name from prior related work is a new approach to attaining sparse solution that allows the natural inclusion of a generative model, relaxation of assumptions on distributions, and ultimately results in more accurate and computationally efficient inference compared to competing algorithms for sparse data decomposition. 

\subsubsection*{Acknowledgments}
I thank Dmitry Malioutov and Rodolfo R'os Zertuche for many insightful discussions and extensive, constructive feedback, as well as David M.~Blei, David Botstein and Kenneth A.~Dawson for support and comments. This work was supported by Irish Research Council for Science, Engineering and Technology (IRCSET) and by Science Foundation Ireland under Grant No.[SFI/SRC/B1155].

\bibliography{RCweb}
\bibliographystyle{msb}

\end{document}